%% file: main.tex
\documentclass{article}

\PassOptionsToPackage{numbers}{natbib}

\usepackage[final]{icbinb2021}

\usepackage[utf8]{inputenc} %
\usepackage[T1]{fontenc}    %
\usepackage{xcolor}         %
\definecolor{hrefcolor}{rgb}{0.0, 0.28, 0.67}
\usepackage{hyperref}       %
\hypersetup{
     colorlinks=true,
     citecolor=hrefcolor,
     linkcolor=hrefcolor,
} %
\usepackage{url}            %
\usepackage{booktabs}       %
\usepackage{amsfonts}       %
\usepackage{nicefrac}       %
\usepackage{microtype}      %
\usepackage{lipsum}
\usepackage{alex-macros} %
\usepackage{caption} %
\usepackage{subcaption}
\usepackage{wrapfig}

\input{macros}
\title{\ourmethod: Categorical probabilistic forecasting with graph structure via local continuous-time dynamics}

\author{%
  Ke Alexander Wang\thanks{Work done as an intern at Amazon Research} \\
  Stanford University \\
  \texttt{alxwang@cs.stanford.edu} \\
  \And
  Danielle Maddix\\
  Amazon Research \\ 
  \texttt{dmmaddix@amazon.com} \\
  \And
  Yuyang Wang \\
  Amazon Research \\ 
  \texttt{yuyawang@amazon.com} \\
}
\begin{document}

\maketitle
\vspace{-.5cm}
\begin{abstract}
We consider the problem of probabilistic forecasting over categories with graph structure, where the dynamics at a vertex depends on its local connectivity structure.
We present \ourmethod, a method that combines the inductive bias of graph neural networks with neural ODEs to capture the intrinsic local continuous-time dynamics of our probabilistic forecasts.
We study the benefits of these two inductive biases by comparing against baseline models that help disentangle the benefits of each.  
We find that capturing the graph structure is crucial for accurate in-domain probabilistic predictions and more sample efficient models. Surprisingly, our experiments demonstrate that the continuous time evolution inductive bias brings little to no benefit despite reflecting the true probability dynamics. 
\end{abstract}

\vspace{-.5cm}
\section{Introduction}

In categorical probabilistic forecasting, we seek to predict a discrete probability distribution \(\pbf(t)\) at some instantaneous time \(t\),  based on observed time-stamped data \citep{Gneiting2014ProbabilisticForecasting}.
Consider the example of forecasting the most likely locations of the next earthquake over a finite set of locations at \(t\), given the history of earthquake times and locations.
We can view locations as vertices \(V\) on a graph \(G=(V,E)\) with edges \(E\) that represent adjacency. 
Specifically, the probability of an earthquake at node \(v\in V\) in the near future is mostly influenced by the probability of earthquakes at nodes within its neighborhood.
This type of graphical structure also appears in other problems, including traffic forecasting \citep{Yu2018SpatiotemporalGraphConvolutional}, information diffusion in social networks \citep{Bakshy2012RoleSocialNetworks}, epidemic diffusion \citep{wang2021, Huang2010DynamicsSISReactiondiffusion}, urban conflict patterns \citep{Linderman2014DiscoveringLatentNetwork}, and is an example of a marked temporal point process \citep{Daley2003IntroductionTheoryPoint}.

In this paper, we consider categorical probabilistic forecasts where there is a graphical structure to inform us of the local dynamics governing \(\pbf(t)\) over time. 
We formalize the intuition that each component of the probability vector \(\pbf(t) \in \reals^{|V|}\) obeys local dynamics using the differential equation
\begin{equation}\label{eqn:local-dynamics}
    \frac{\dd p_v}{\dd t}  = g\parens*{p_v, \{p_u \mid u\in N(v)\}, t},
\end{equation}
which we use to inform our model's inductive bias.
Here, \(g\) governs the local dynamics,  \(\Ncal(v) \subseteq V\) denotes the set of neighboring nodes of \(v\), and \(p_v\) denotes the probability at node \(v\).
To capture the equivariant local dynamics of our forecast \(\pbf(t)\), we propose \ourmethod, a model that learns a neural ODE \citep{Chen2018NeuralOrdinaryDifferential} with graph neural network (GNN) \citep{Wu2021ComprehensiveSurveyGraph} dynamics.

Our method \ourmethod\  introduces two inductive biases to aid with probabilistic forecasting over graph-structured categories by 1) utilizing graph structure explicitly and 2) introducing temporal evolution through a neural ODE. 
To disentangle the benefits of these two biases, we introduce two baseline models, ablating each bias.
We find that utilizing the known graph structure results is key, and results in 10x improvements in accuracy and sample efficiency.
On the other hand, explicitly modelling the temporal dynamics surprisingly results in little benefits.

\section{\ourmethod: Forecasting with temporal dynamics and graph structure}
Let \(G=(V,E)\) be a graph, and let \(t_i\in \reals_0^+\) denote the timestamp of an event at node \(v_i \in V\).
Given \(G\) and an irregularly sampled dataset \(\Dcal = \{(t_i, v_i)\}_{i=1}^N\), we want to learn the probability vector \(\pbf(t) \in \reals^{|V|}\) of each \(v\in V\) at any time \(t\). %
We wish to model the dynamics of \(\pbf(t)\) such that the change in the probability at node \(v\) depends only on the neighborhood \(\Ncal(v)\) around \(v\), as described in \autoref{eqn:local-dynamics}.
However, directly parameterizing \(g\) from \autoref{eqn:local-dynamics} with a neural ODE can violate conservation of probability: \(\onevec\transpose \pbf(t) = 1\).

Instead of explicitly enforcing the sum constraint into our neural ODE, we model the dynamics in a continuous-time embedding space from which we derive the dynamics \(\dd \pbf/ \dd t\).  
Specifically, let \(\Zbf_0 \in \reals^{|V|\times D}\), where \(D\) denotes the embedding space dimension.
We use \(\zvec_{0,i}\) to denote row \(i\) of \(\Zbf_0\) at initial time \(t_0\), corresponding to the embedding of node \(v_i\). 
We then model the dynamics of the continuous-time embeddings \(\Zbf(t) \in \reals^{|V|\times D}\) via 
\begin{align}\label{eqn:embedding-dynamics}
    \frac{\dd \Zbf}{\dd t} &= g(\Zbf, G, t) \quad \subjectto \quad \Zbf(t_0) = \Zbf_0,
\end{align}
where \(g\) is the learned graph neural network (GNN) dynamics.
To map \(\Zbf(t)\) to a probability space while preserving equivariance, we learn a shared projection \(\pi: \reals^{D} \to \reals\) such that
\begin{align}
    \pbf(t) = \Softmax{\pi(\zvec(t)_1), \ldots, \pi(\zvec(t)_{|V|})}.
\end{align}
Provided that \(g\) and \(\pi\) are differentiable, which can be done using smooth activation functions, our model then implicitly models the local temporal dynamic of our problem in \autoref{eqn:local-dynamics}.
Finally, we train our model \ourmethod\ by maximizing the log likelihood \(\sum_{i=1}^N \log p_{v_i}(t_i)\) with respect to the parameters of \(\pi\), \(g\), and the initial condition \(\Zbf_0\).

\paragraph{Incorporating node attributes.}
In some cases \(G\) may have node attributes \(\{a_v\}_{v\in V}\) for each node \(v\in V\) that affect the interaction dynamics, such as the geographical coordinates of each node in a spatial graph or the demographics of a user in a social network.
Node attributes can be easily incorporated by letting the initial node embeddings be a learned function of the attributes, \(\Zbf_0[v] = \psi_v(a_v)\), and optimizing with respect to the parameters of \(\{\psi_v\}\).

\paragraph{Related works.}
Our paper lies at the intersection of probabilistic forecasting, neural ODEs, and graph neural networks (GNNS), and can be seen as the discrete analogue of continuous normalizing flows \citep{Chen2018NeuralOrdinaryDifferential,Grathwohl2018FFJORDFreeFormContinuous, Chen2020NeuralSpatioTemporalPoint} on manifolds \citep{Lou2020NeuralManifoldOrdinary, Mathieu2020RiemannianContinuousNormalizing}.
Probabilistic forecasting seeks to predict a full distribution at each time step \citep{hyndman2018,Gneiting2014ProbabilisticForecasting}, with contemporary methods often relying on deep probabilistic models \citep{Salinas2020DeepARProbabilisticForecasting, Wang2019DeepFactorsForecasting, Rasul2020MultivariateProbabilisticTime, Rangapuram2018DeepStateSpace}.
A direct application of categorical probabilistic forecasts is marked temporal point processes, which learn the rate of an event type \(v\) at time \(t\), summarized by the conditional intensity function \(\lambda(t,v) = \lambda(t)\cdot p_v(t)\) \citep{Daley2003IntroductionTheoryPoint}.
The inductive bias of a learnable ODE with GNN dynamics has also been explored in the context of other problems, including graph generation \citep{Deng2019ContinuousGraphFlow}, node classification \citep{Poli2021ContinuousDepthNeuralModels, Chamberlain2021GRANDGraphNeural}, multi-particle trajectory prediction \citep{Poli2021ContinuousDepthNeuralModels}, learning partial differential equations \citep{Iakovlev2020LearningContinuoustimePDEs}, and knowledge graph forecasting \citep{Han2021TemporalKnowledgeGraph}. 

\section{Results: I Can't Believe Temporal Dynamics Don't Matter!}
\paragraph{Synthetic datasets.}
\begin{figure}
    \centering
    \begin{minipage}{\linewidth}
        \includegraphics[width=\linewidth]{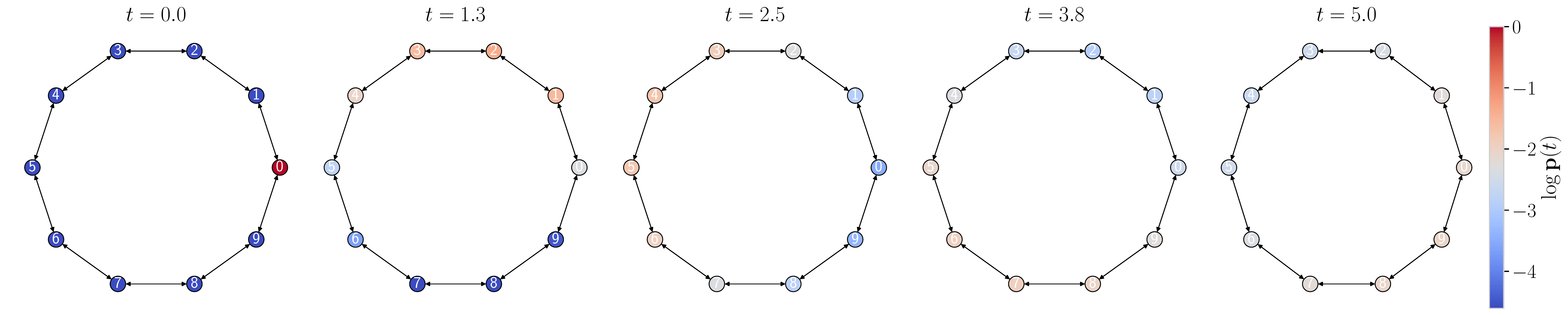}
    \end{minipage} \hfill
    \begin{minipage}{\linewidth}
        \includegraphics[width=\linewidth]{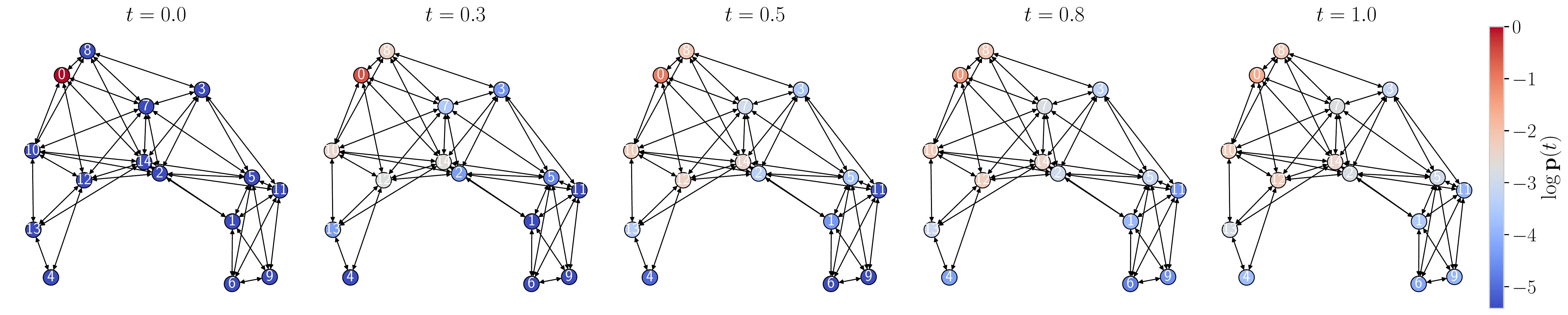}
    \end{minipage} \hfill
    \caption{(Top) Advection on a cyclic graph set chosen so that the probability mass is transported more strongly in the counter-clockwise direction. (Bottom) Advection on a random geometric graph with randomly chosen edge weights. (Both) Colors are shown in log-scale to make dynamics more visually-apparent. Light-gray coloring corresponds to \(\log (1 / |V|)\) which is the steady-state probability mass for each node.}
    \label{fig:advection}
\end{figure}
We apply our method to model the mark component of a marked temporal point process (TPP) occurring on the nodes of a graph such that \(p_v(t)\) is the probability of an event occurring on vertex \(v\) at time \(t\).
We create a synthetic dataset where events occur over time on a directed graph \(G\), with node probabilities that obey graph advection as an example of local dynamics \citep{chapman2011}.
Graph advection conserves the total probability by ensuring \(\onevec\transpose \dd \pbf/\dd t = 0\).
We represent the graph \(G\) by the weighted adjacency matrix \(A\), where \(A_{uv} > 0\) for \((u,v) \in E\).
We sample sequences of events over time \([0, T]\) from a homogeneous Poisson process with constant temporal intensity \(\lambda(t) = \lambda = 2.5\) and temporal node probability \(\pbf(t) \in \reals^{|V|}\) governed by the graph advection equation \citep{chapman2011}
\begin{equation}
    \frac{\dd \pbf}{\dd t} = -\Lout(A)\transpose \pbf \iff \frac{\dd p_v}{\dd t} = \sum_{v:\ (v,u)\in E} A_{vu}p_v - \sum_{v:\ (u, v)\in E} A_{uv}p_u.
\end{equation}
Here, \(\Lout(A):= \Dout(A) - A\) denotes the out-degree graph Laplacian, and \(\Dout(G)\) denotes the diagonal out-degree matrix with \(\Dout(G)_{ii} = \sum_j A_{ij}\).

We create two graphs structures for our synthetic datasets, a ring graph and a random geometric graph, and visualize their advection on the graph over time in \autoref{fig:advection}; see \autoref{sec:implementation-details} for more details on their construction.
We also visualize the advection dynamics of each component of \(\pbf(t)\) for the ring graph in \autoref{fig:advection-ring} of \autoref{sec:implementation-details}. 
We use \(T=5\) seconds for the ring graph dataset and \(T=1\) second for the geometric graph dataset.
Since the timestamps are sampled from a Poisson process and \emph{not} equidistantly spaced over \([0, T]\), the continuous time aspect of the problem is clearly evident in the dataset.

\paragraph{Evaluating each inductive bias of \ourmethod.}
\begin{wrapfigure}{r}{0.5\textwidth}
  \vspace{-5mm}
  \begin{center}
    \includegraphics[width=0.5\textwidth]{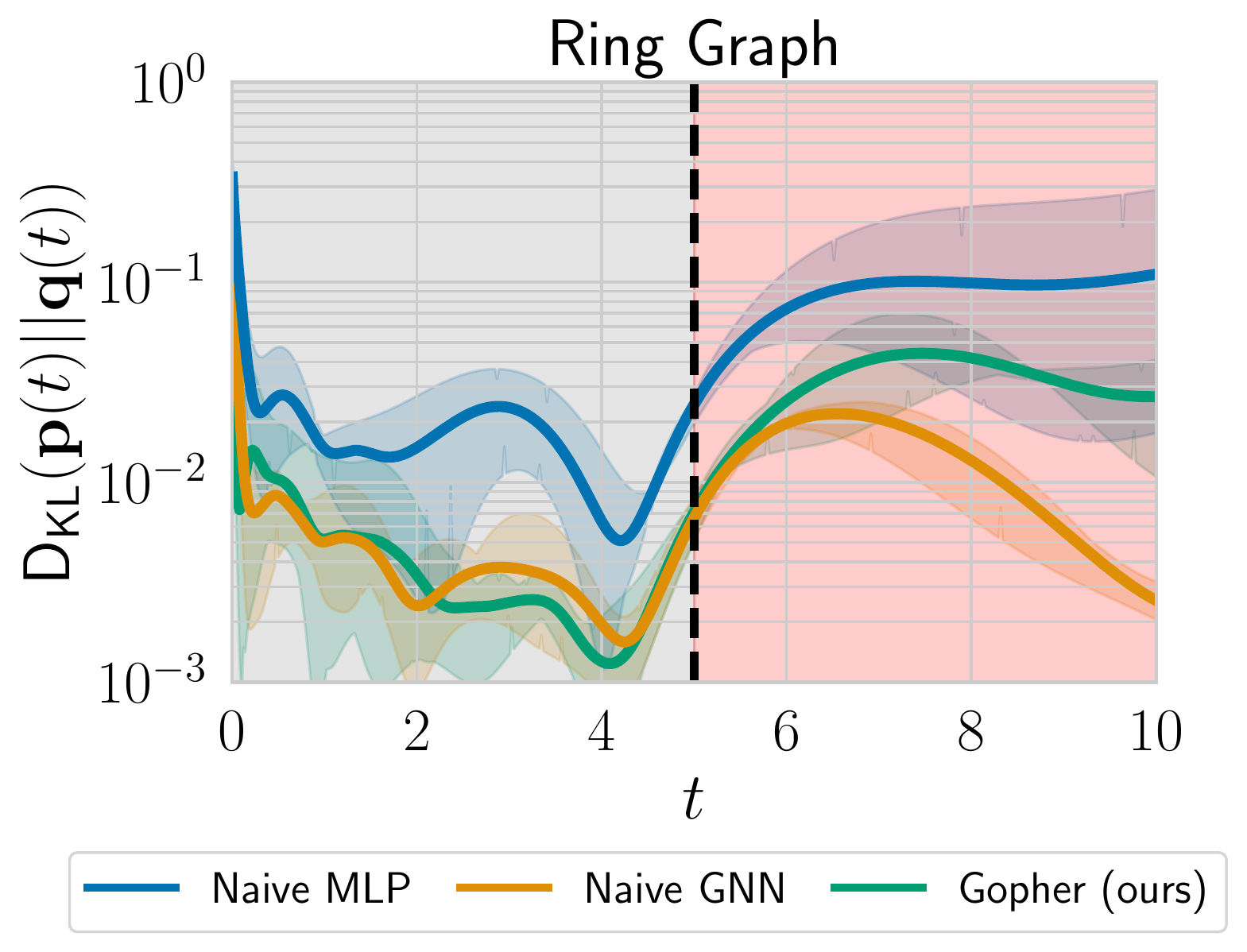}
  \end{center}
  \caption{KL divergence between the true probabilities \(\pbf(t)\) and the predicted probabilities \(\qbf(t)\) trained on 1024 sampled sequences. Gray is the training set time interval \([0,T]\) and red is the extrapolation region \([T, 2T]\) beyond the training set.}
  \label{fig:kl-divergence}
  \vspace{-2mm}
\end{wrapfigure}
We evaluate the accuracy and sample-efficiency improvements from incorporating graph structure and modelling temporal dynamics in \ourmethod.
To disentangle the effects of these two inductive biases, we compare our model to two baseline models.
The first is a two-layer MLP that acts on node-embeddings concatenated with time, which has none of the above inductive biases.
The second is a single-layer GNN that also acts on node-embeddings concatenated with time. The GNN incorporates the explicit graph structure, but does not incorporate dynamical systems structure.
We refer to the models as \naivemlp\, and \naivegnn\,  respectively.
In our experiments, \ourmethod\, learns \(g\) using a Graph Isomorphism Network (GIN) layer \citep{Xu2018HowPowerfulAre} parameterized by a two-layer MLP; we use another two-layer MLP for the projection \(\pi\).
\naivegnn\, uses the same GIN architecture and projection except that it does not learn a differential equation.
Finally, \naivemlp\, replaces the GIN layer with a two layer MLP.
See \autoref{sec:implementation-details} for further details on our experiment hyperparameters.

\begin{figure}
    \vspace{-5mm}
    \centering
    \includegraphics[width=\linewidth]{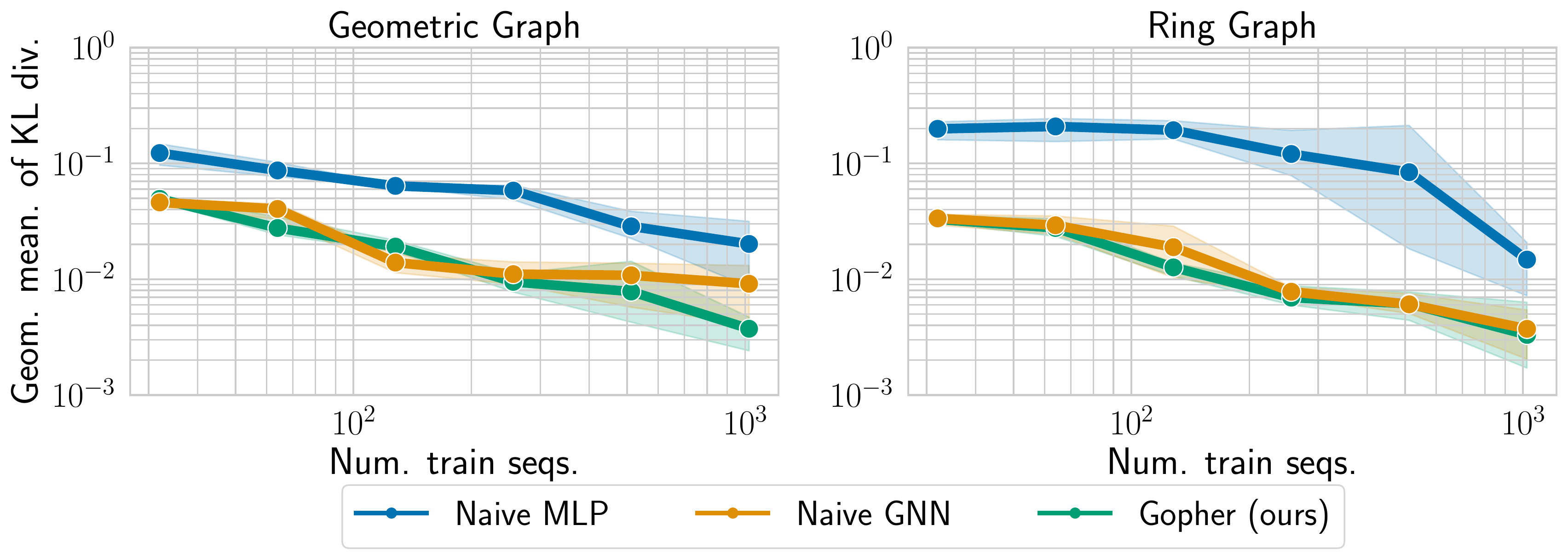}
    \caption{Sample complexity of each model on the geometric graph dataset (left) and cyclic graph dataset (right). The error is measured in terms of the geometric mean of the KL divergence over the evaluation time period \([0, T]\). Prediction of uniform probability corresponds to a geometric mean KL divergence of \(0.72\) for the geometric graph and \(0.15\) for the cyclic graph.}
    \label{fig:sample-complexity}
\end{figure}

\autoref{fig:kl-divergence} shows the KL divergence betweeen the ground truth \(\pbf(t)\) and the learned predictions over time for the ring graph.
We summarize the KL divergence over \([0, T]\) in \autoref{fig:sample-complexity} by the geometric mean since the error varies over multiple overs of magnitudes over time \citep{Finzi2020SimplifyingHamiltonianLagrangian}.
In both figures, we show the 95\% confidence intervals over 3 seeds.
For both datasets, there is 10x difference in accuracy between the graph structured models and \naivemlp, indicating that utilizing the graph structure is greatly beneficial.
Though \naivegnn\, does not explicitly model the local temporal dynamics of the datasets, it performs nearly identically to our model \ourmethod\, in fitting \(\pbf(t)\) over the training interval \([0, T]\).
In principle, \ourmethod\, has the best chance of extrapolating to the \([T, 2T]\) time period not seen during training since \ourmethod\, explicitly models the local dynamics.
However, \ourmethod's\, poor extrapolation ability suggests that its learned dynamics do not actually reflect the true dynamics.
Indeed, in \autoref{fig:edge-deletion} of \autoref{sec:model-robustness} we show that although \ourmethod\, can fit the training data well, it is brittle to edge deletions, further indicating \ourmethod\, does not learn the true dynamics.

\paragraph{Real-world dataset.}
\begin{figure}[H]
    \vspace{-5mm}
    \centering
    \includegraphics[width=\linewidth]{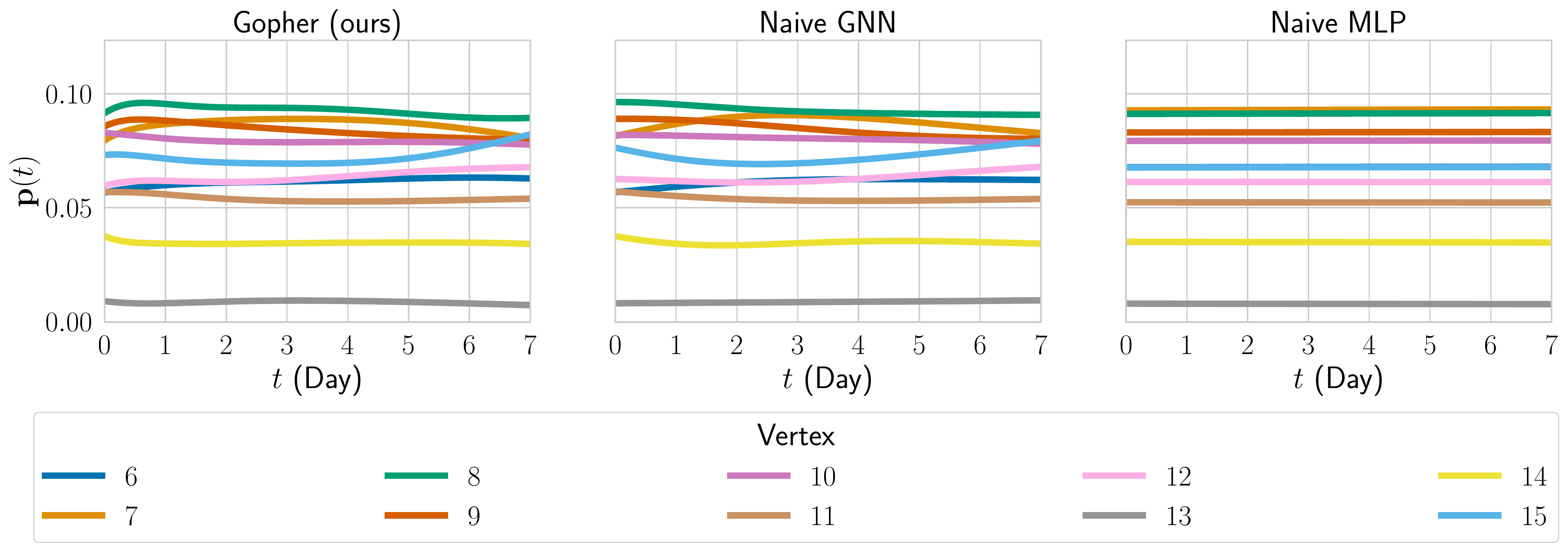}
    \caption{Learned \(\pbf(t)\) over a subset of the counties fit using the COVID-19 dataset preprocessed by \citet{Chen2020NeuralSpatioTemporalPoint}. Only the graph-based models are able to capture the variations over time. See \autoref{sec:distribution-shift} for the empirical distribution of \(\pbf(t)\).}
    \label{fig:covid-vertex-distribution}
    \vspace{-5mm}
\end{figure}
We use data released publicly by the New York Times \cite{NewYorkTimes2021} on daily COVID-19 cases in New Jersey state to construct a real-world categorical probabilistic forecasting dataset, following the preprocessing script of \citet{Chen2020NeuralSpatioTemporalPoint}. 
We aggregate the cases by county and form a graph with 21 nodes where each node is a county and each edge is a county border.
Using the train/test split from \citet{Chen2020NeuralSpatioTemporalPoint}, we obtain per event log likelihoods with 1-standard-deviations of  \(-2.766 \pm 0.003\) for \naivemlp,\ \(-2.768 \pm 0.003\) for \naivegnn,\ and \(-2.767 \pm 0.000\) for \ourmethod\ over 3 seeds.
However, these likelihoods are not representative of the model differences since we find a large distribution shift between the train and test distribution shown in \autoref{fig:distribution-shift} of \autoref{sec:distribution-shift}. This distribution shift causes the models to perform equally poorly on the test set.
In actuality, \naivemlp\, completely fails to capture variations in \(\pbf(t)\) over time, as shown in \autoref{fig:covid-vertex-distribution}.

\section{Discussion}
Although the inductive biases of \ourmethod, directly reflect properties of categorical forecasting with local continuous-time dynamics, our experiments find that, surprisingly, explicitly modelling the temporal dynamics does not improve performance.
Most of the performance gains of \ourmethod\, come from incorporating a graph structure, which can be done with a simple baseline model like \naivegnn.\ 
The failure of \ourmethod\, can be attributed to the fact that the learned dynamics in the embedding space do not accurately reflect the ground truth dynamics in probability space. %

{
\small
\bibliographystyle{plainnat}
\bibliography{references}
\clearpage
}

\appendix
\section{Distribution shift in our real-world dataset}\label{sec:distribution-shift}
\begin{figure}[H]
    \centering
    \includegraphics[width=\linewidth]{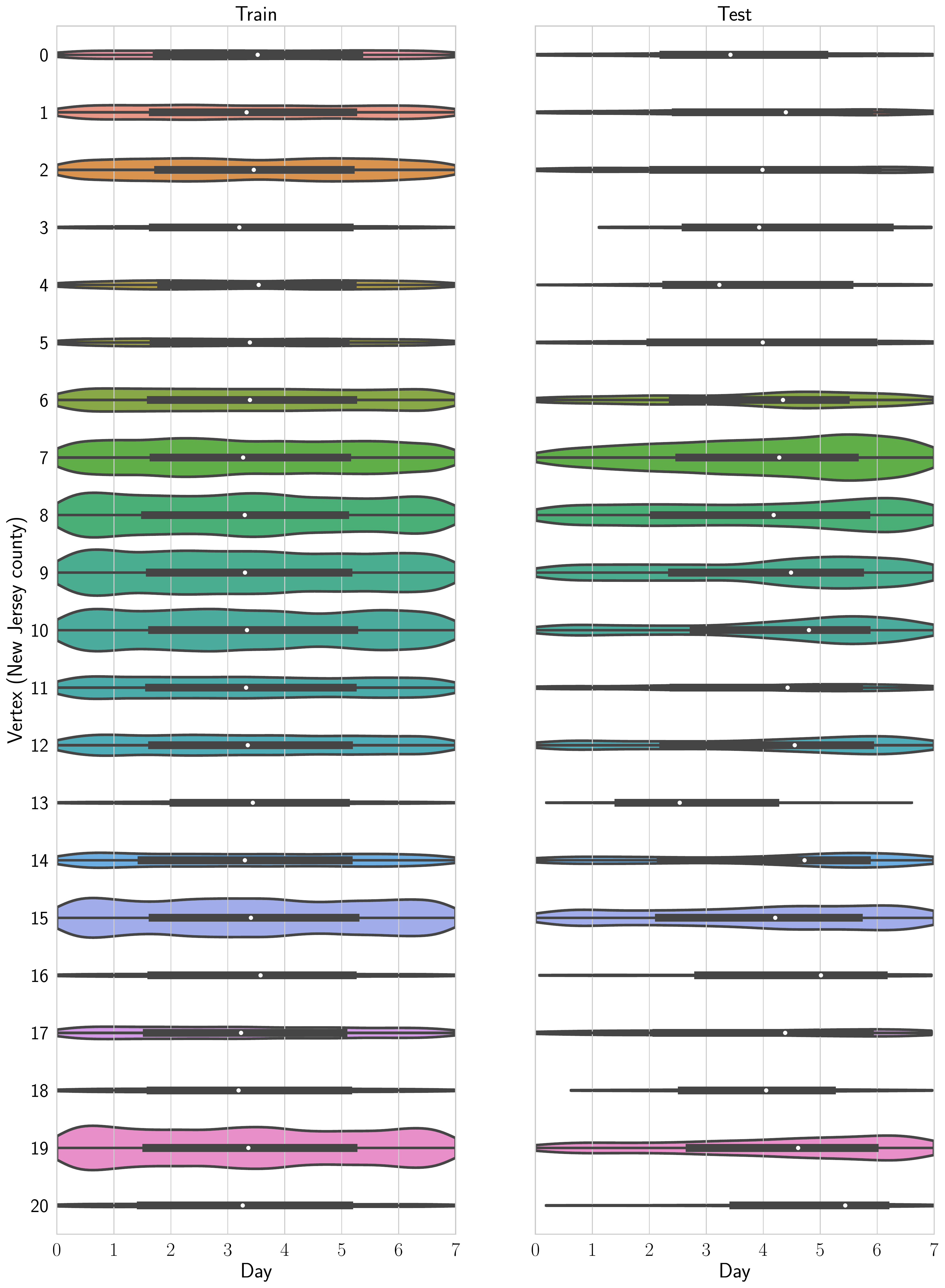}
    \caption{Distribution shift between the train and test distribution of the New Jersey COVID-19 cases created by \citet{Chen2020NeuralSpatioTemporalPoint} when binned by the 21 counties. For each of the two distributions, the height of each violin plots is normalized by the total count of observations in that split, i.e. size of training set or size of test set. For most vertices, there are fewer COVID cases later in the 7 day interval in the test set than in the training set.}
    \label{fig:distribution-shift}
\end{figure}

\section{Learned forecasts on COVID-19 dataset}
\begin{figure}[H]
    \centering
    \includegraphics[width=\linewidth]{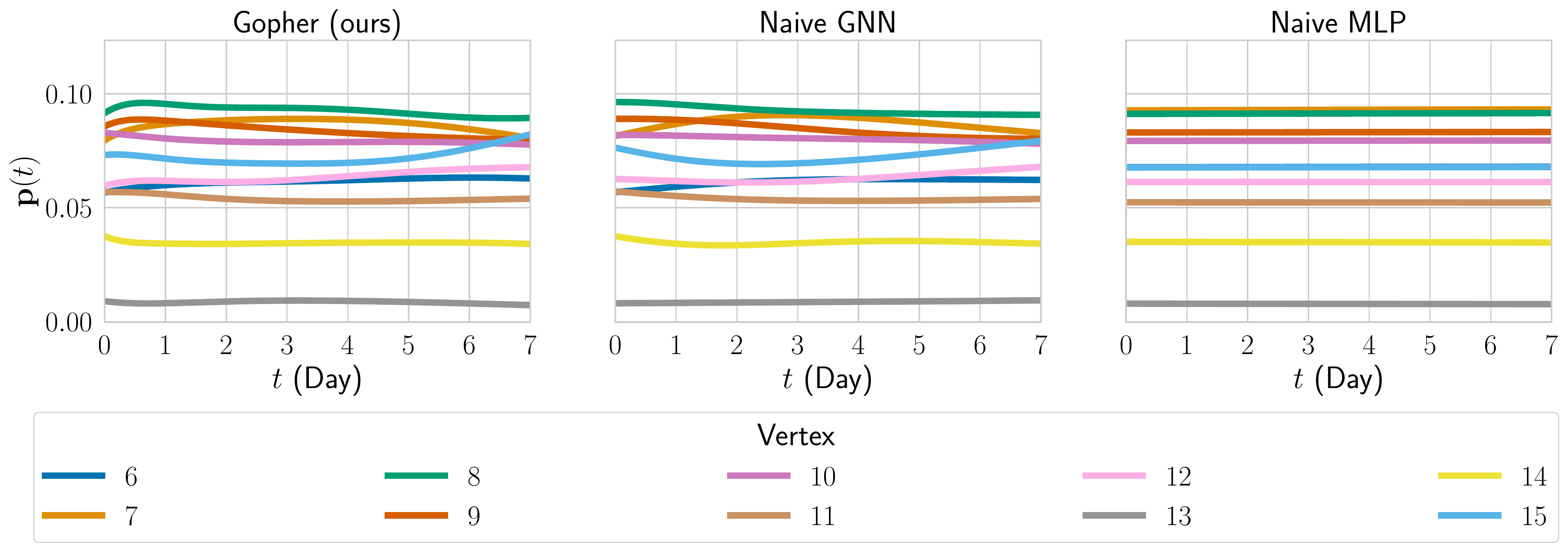}
    \caption{A copy of \autoref{fig:covid-vertex-distribution} for easier comparison to the empirical distribution in \autoref{fig:distribution-shift}}
    \label{fig:covid-vertex-distribution-2}
\end{figure}

\section{Model robustness}\label{sec:model-robustness}
\begin{figure}[H]
    \centering
    \begin{minipage}{\linewidth}
        \includegraphics[width=\linewidth]{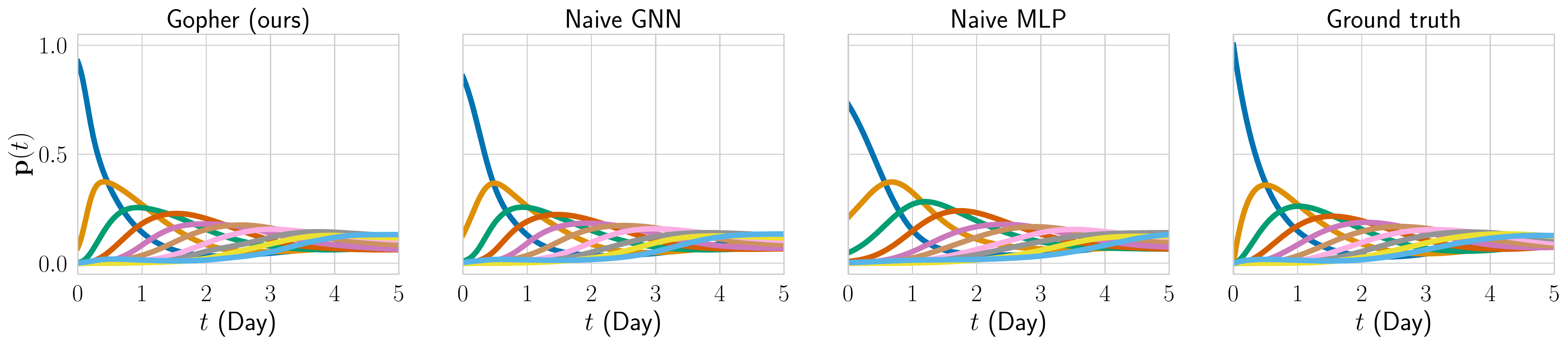}
    \end{minipage} \hfill
    \begin{minipage}{\linewidth}
        \includegraphics[width=\linewidth]{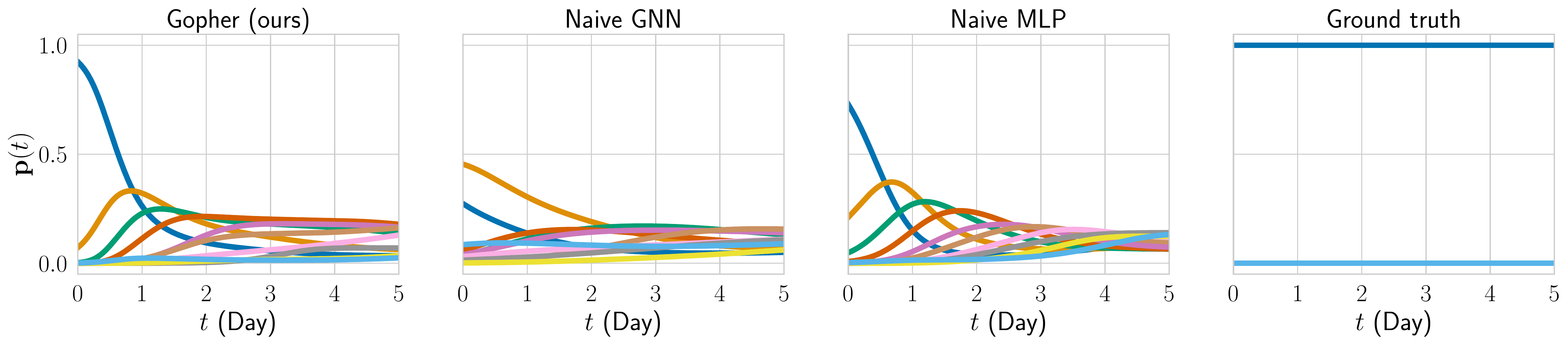}
    \end{minipage} \hfill
    \caption{(Top) The learned and ground truth \(\pbf(t)\) for the ring graph dataset. (Bottom) The predicted \(\pbf(t)\) after we remove all edges from the ring graph. Notice that the ground truth of a completely disconnected graph is to have \(\pbf(t) = \pbf(0)\) for all \(t\). However, all of the models fail completely on this new disconnected graph, suggesting that they do not learn the true dependence of \(\pbf(t)\) on the graph structure.}
    \label{fig:edge-deletion}
\end{figure}

\section{Implementation details}\label{sec:implementation-details}
\begin{figure}
    \centering
    \includegraphics[width=\linewidth]{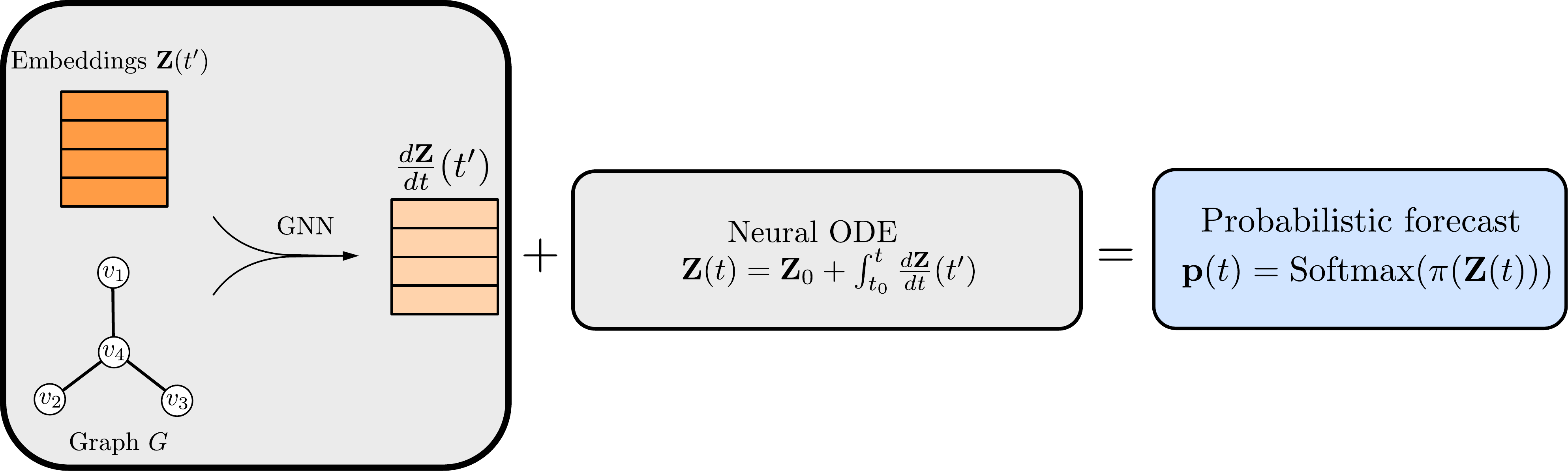}
    \caption{Architecture of \ourmethod.}
    \label{fig:gopher}
\end{figure}
We use 2 layer MLPs with 64 hidden units per layer whenever we use a MLP. We use Swish activations to ensure smoothness of our dynamics.
We also use an augmented neural ODE \citep{Dupont2019AugmentedNeuralODEs}, using 16 dimensions as augmented dimensions out of the 64 hidden dimensions.

\subsection{Model architectures.}
\paragraph{\ourmethod.} We parameterize the dynamics \(g\) from \autoref{eqn:embedding-dynamics} using one graph isomorophism network layer parameterized by a MLP. 
We also use a MLP to model the projection \(\pi\).

\paragraph{\naivegnn.} We use the same architecture as \ourmethod,\, namely a GIN layer followed by a projection \(\pi\).\, However, instead of using the model to parameterize ODE dynamics, we directly input the node embeddings concatenated with time \(t\) through the GNN.

\paragraph{\naivemlp.} We replace the GIN layer of \naivegnn\, with a MLP, keeping all else the same.

\subsection{Training procedure and dataset details.}
To maximize hardware parallelism, we parallelize our neural ODE computation across sequence timesteps and across sequences using the time-reparameterization trick outlined in \citet{Chen2020NeuralSpatioTemporalPoint}.
For the synthetic datasets, we use the AdamW optimizer with 0.01 learning rate and batch size 64 for 30 epochs.
For the COVID-19 dataset, we use a \(3\times 10^{-4}\) learning rate and batch size 4 for 15 epochs.
Here, each batch consists of multiple sequences drawn from the training period \([0, T]\).  
We use \(T=5\) for the ring graph, \(T=1\) for the geometric graph, and \(T=8\) for the New Jersey counties graph.
We generate the ring graph dataset by using hand-set coefficients for the edge weights \(A_{uv}\) to allow for counter-clockwise transport.
We generate the geometric graph dataset by generating a random geometric graph via the \verb|networkx| python package and drawing a random sample of \(\{A_{uv}\}\).
\begin{figure}[H]
  \vspace{-5mm}
  \begin{center}
    \includegraphics[width=0.5\textwidth]{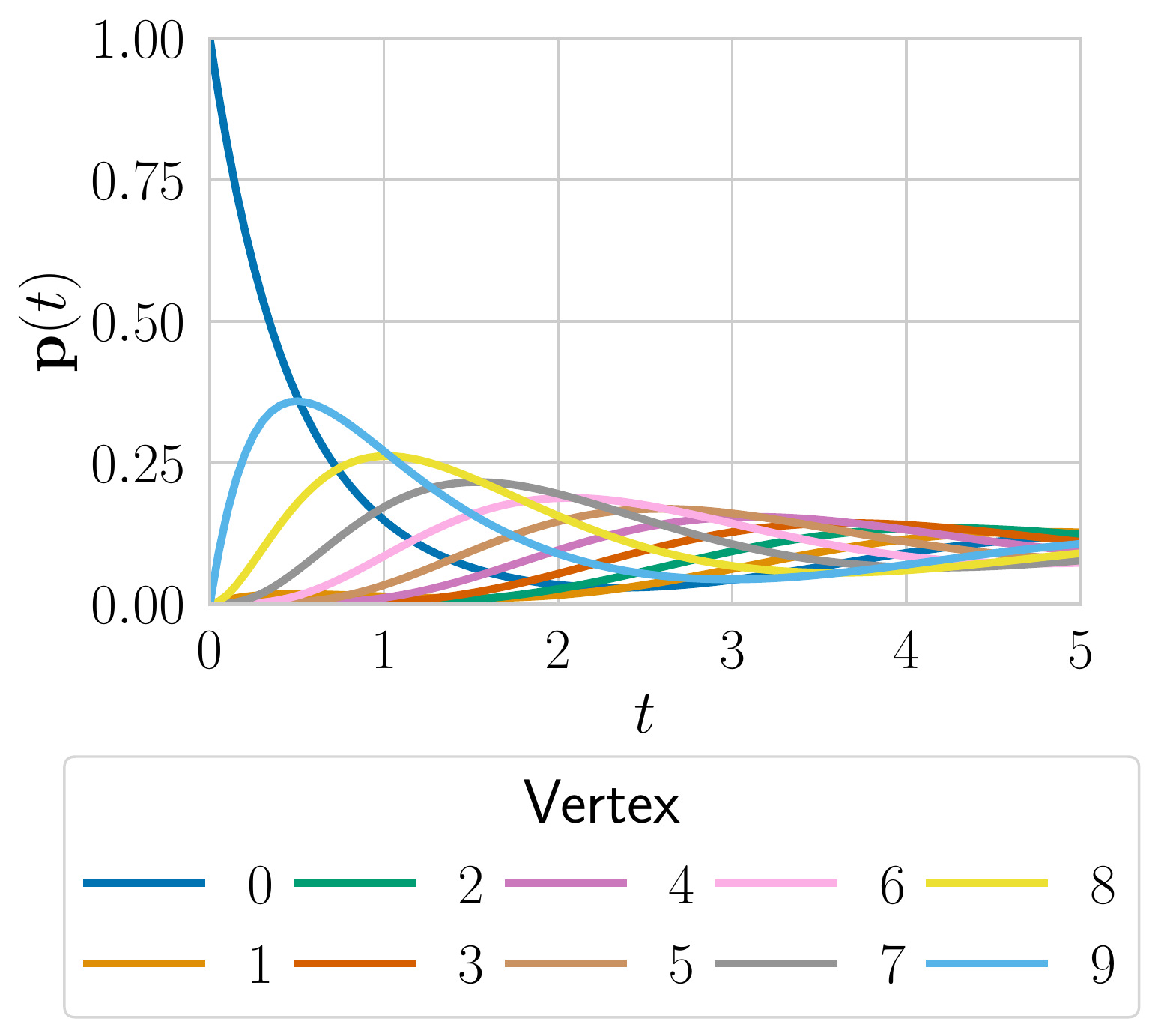}
  \end{center}
  \caption{The dynamics of each component of \(\pbf(t)\) on the cyclic graph. Each component corresponds to the probability of a vertex on the graph.} %
  \label{fig:advection-ring}
\end{figure}

\end{document}

%% file: macros.tex
\newcommand*{\Lout}{L_{\text{out}}}
\newcommand*{\Dout}{D_{\text{out}}}
\newcommand*{\naivemlp}{\textsc{NaiveMLP}}
\newcommand*{\naivegnn}{\textsc{NaiveGNN}}

\newcommand*{\Softmax}[1]{\operatorname{Softmax}\parens*{#1}}
\newcommand*{\ourmethod}{\textsc{Gopher}}

\definecolor{mplblue}{HTML}{0173B2}
\definecolor{mplorange}{HTML}{DE8F05}
\definecolor{mplgreen}{HTML}{029E73}
\definecolor{mplred}{HTML}{F17B7B}

%% file: main.bbl
\begin{thebibliography}{28}
\providecommand{\natexlab}[1]{#1}
\providecommand{\url}[1]{\texttt{#1}}
\expandafter\ifx\csname urlstyle\endcsname\relax
  \providecommand{\doi}[1]{doi: #1}\else
  \providecommand{\doi}{doi: \begingroup \urlstyle{rm}\Url}\fi

\bibitem[Bakshy et~al.(2012)Bakshy, Rosenn, Marlow, and
  Adamic]{Bakshy2012RoleSocialNetworks}
Eytan Bakshy, Itamar Rosenn, Cameron Marlow, and Lada Adamic.
\newblock The role of social networks in information diffusion.
\newblock In \emph{Proceedings of the 21st International Conference on {{World
  Wide Web}}}, {{WWW}} '12, pages 519--528, {New York, NY, USA}, April 2012.
  {Association for Computing Machinery}.
\newblock ISBN 978-1-4503-1229-5.
\newblock \doi{10.1145/2187836.2187907}.

\bibitem[Chamberlain et~al.(2021)Chamberlain, Rowbottom, Gorinova, Bronstein,
  Webb, and Rossi]{Chamberlain2021GRANDGraphNeural}
Ben Chamberlain, James Rowbottom, Maria~I. Gorinova, Michael Bronstein, Stefan
  Webb, and Emanuele Rossi.
\newblock {{GRAND}}: {{Graph Neural Diffusion}}.
\newblock In \emph{International {{Conference}} on {{Machine Learning}}}, pages
  1407--1418. {PMLR}, July 2021.

\bibitem[Chapman and Mesbahi(2011)]{chapman2011}
Airlie Chapman and Mehran Mesbahi.
\newblock Advection on graphs.
\newblock \emph{IEEE Conference on Decision and Control and European Control
  Confereence (CDC-ECC)}, 50:\penalty0 1461--1466, December 2011.

\bibitem[Chen et~al.(2018)Chen, Rubanova, Bettencourt, and
  Duvenaud]{Chen2018NeuralOrdinaryDifferential}
Ricky T.~Q. Chen, Yulia Rubanova, Jesse Bettencourt, and David~K Duvenaud.
\newblock Neural {{Ordinary Differential Equations}}.
\newblock In \emph{Advances in {{Neural Information Processing Systems}}},
  volume~31. {Curran Associates, Inc.}, 2018.

\bibitem[Chen et~al.(2020)Chen, Amos, and
  Nickel]{Chen2020NeuralSpatioTemporalPoint}
Ricky T.~Q. Chen, Brandon Amos, and Maximilian Nickel.
\newblock Neural {{Spatio}}-{{Temporal Point Processes}}.
\newblock In \emph{International {{Conference}} on {{Learning
  Representations}}}, September 2020.

\bibitem[Daley and {Vere-Jones}(2003)]{Daley2003IntroductionTheoryPoint}
D.~J. Daley and D.~{Vere-Jones}.
\newblock \emph{An {{Introduction}} to the {{Theory}} of {{Point Processes}}:
  {{Volume I}}: {{Elementary Theory}} and {{Methods}}}.
\newblock Probability and {{Its Applications}}, {{An Introduction}} to the
  {{Theory}} of {{Point Processes}}. {Springer-Verlag}, {New York}, second
  edition, 2003.
\newblock ISBN 978-0-387-95541-4.
\newblock \doi{10.1007/b97277}.

\bibitem[Deng et~al.(2019)Deng, Nawhal, Meng, and
  Mori]{Deng2019ContinuousGraphFlow}
Zhiwei Deng, Megha Nawhal, Lili Meng, and Greg Mori.
\newblock Continuous {{Graph Flow}}.
\newblock \emph{arXiv:1908.02436 [cs, stat]}, September 2019.

\bibitem[Dupont et~al.(2019)Dupont, Doucet, and
  Teh]{Dupont2019AugmentedNeuralODEs}
Emilien Dupont, Arnaud Doucet, and Yee~Whye Teh.
\newblock Augmented {{Neural ODEs}}.
\newblock In \emph{Advances in {{Neural Information Processing Systems}}},
  volume~32. {Curran Associates, Inc.}, 2019.

\bibitem[Finzi et~al.(2020)Finzi, Wang, and
  Wilson]{Finzi2020SimplifyingHamiltonianLagrangian}
Marc Finzi, Ke~Alexander Wang, and Andrew~G. Wilson.
\newblock Simplifying {{Hamiltonian}} and {{Lagrangian Neural Networks}} via
  {{Explicit Constraints}}.
\newblock \emph{Advances in Neural Information Processing Systems},
  33:\penalty0 13880--13889, 2020.

\bibitem[Gneiting and Katzfuss(2014)]{Gneiting2014ProbabilisticForecasting}
Tilmann Gneiting and Matthias Katzfuss.
\newblock Probabilistic {{Forecasting}}.
\newblock \emph{Annual Review of Statistics and Its Application}, 1\penalty0
  (1):\penalty0 125--151, 2014.
\newblock \doi{10.1146/annurev-statistics-062713-085831}.

\bibitem[Grathwohl et~al.(2018)Grathwohl, Chen, Bettencourt, Sutskever, and
  Duvenaud]{Grathwohl2018FFJORDFreeFormContinuous}
Will Grathwohl, Ricky T.~Q. Chen, Jesse Bettencourt, Ilya Sutskever, and David
  Duvenaud.
\newblock {{FFJORD}}: {{Free}}-{{Form Continuous Dynamics}} for {{Scalable
  Reversible Generative Models}}.
\newblock In \emph{International {{Conference}} on {{Learning
  Representations}}}, September 2018.

\bibitem[Han et~al.(2021)Han, Ding, Ma, Gu, and
  Tresp]{Han2021TemporalKnowledgeGraph}
Zhen Han, Zifeng Ding, Yunpu Ma, Yujia Gu, and Volker Tresp.
\newblock Temporal {{Knowledge Graph Forecasting}} with {{Neural ODE}}.
\newblock \emph{arXiv:2101.05151 [cs]}, August 2021.

\bibitem[Huang et~al.(2010)Huang, Han, and
  Liu]{Huang2010DynamicsSISReactiondiffusion}
Wenzhang Huang, Maoan Han, and Kaiyu Liu.
\newblock Dynamics of an {{SIS}} reaction-diffusion epidemic model for disease
  transmission.
\newblock \emph{Mathematical Biosciences \& Engineering}, 7\penalty0
  (1):\penalty0 51, 2010.
\newblock \doi{10.3934/mbe.2010.7.51}.

\bibitem[Hyndman and Athanasopoulos(2018)]{hyndman2018}
{Robin John} Hyndman and George Athanasopoulos.
\newblock \emph{Forecasting: Principles and Practice}.
\newblock OTexts, Australia, 2nd edition, 2018.

\bibitem[Iakovlev et~al.(2020)Iakovlev, Heinonen, and
  L{\"a}hdesm{\"a}ki]{Iakovlev2020LearningContinuoustimePDEs}
Valerii Iakovlev, Markus Heinonen, and Harri L{\"a}hdesm{\"a}ki.
\newblock Learning continuous-time {{PDEs}} from sparse data with graph neural
  networks.
\newblock In \emph{International {{Conference}} on {{Learning
  Representations}}}, September 2020.

\bibitem[Linderman and Adams(2014)]{Linderman2014DiscoveringLatentNetwork}
Scott Linderman and Ryan Adams.
\newblock Discovering {{Latent Network Structure}} in {{Point Process Data}}.
\newblock In \emph{International {{Conference}} on {{Machine Learning}}}, pages
  1413--1421. {PMLR}, June 2014.

\bibitem[Lou et~al.(2020)Lou, Lim, Katsman, Huang, Jiang, Lim, and
  De~Sa]{Lou2020NeuralManifoldOrdinary}
Aaron Lou, Derek Lim, Isay Katsman, Leo Huang, Qingxuan Jiang, Ser~Nam Lim, and
  Christopher~M De~Sa.
\newblock Neural manifold ordinary differential equations.
\newblock In H.~Larochelle, M.~Ranzato, R.~Hadsell, M.~F. Balcan, and H.~Lin,
  editors, \emph{Advances in Neural Information Processing Systems}, volume~33,
  pages 17548--17558. {Curran Associates, Inc.}, 2020.

\bibitem[Mathieu and Nickel(2020)]{Mathieu2020RiemannianContinuousNormalizing}
Emile Mathieu and Maximilian Nickel.
\newblock Riemannian {{Continuous Normalizing Flows}}.
\newblock In H.~Larochelle, M.~Ranzato, R.~Hadsell, M.~F. Balcan, and H.~Lin,
  editors, \emph{Advances in {{Neural Information Processing Systems}}},
  volume~33, pages 2503--2515. {Curran Associates, Inc.}, 2020.

\bibitem[Poli et~al.(2021)Poli, Massaroli, Rabideau, Park, Yamashita, Asama,
  and Park]{Poli2021ContinuousDepthNeuralModels}
Michael Poli, Stefano Massaroli, Clayton~M. Rabideau, Junyoung Park, Atsushi
  Yamashita, Hajime Asama, and Jinkyoo Park.
\newblock Continuous-{{Depth Neural Models}} for {{Dynamic Graph Prediction}}.
\newblock \emph{arXiv:2106.11581 [cs, stat]}, June 2021.

\bibitem[Rangapuram et~al.(2018)Rangapuram, Seeger, Gasthaus, Stella, Wang, and
  Januschowski]{Rangapuram2018DeepStateSpace}
Syama~Sundar Rangapuram, Matthias~W Seeger, Jan Gasthaus, Lorenzo Stella,
  Yuyang Wang, and Tim Januschowski.
\newblock Deep {{State Space Models}} for {{Time Series Forecasting}}.
\newblock In \emph{Advances in {{Neural Information Processing Systems}}},
  volume~31. {Curran Associates, Inc.}, 2018.

\bibitem[Rasul et~al.(2020)Rasul, Sheikh, Schuster, Bergmann, and
  Vollgraf]{Rasul2020MultivariateProbabilisticTime}
Kashif Rasul, Abdul-Saboor Sheikh, Ingmar Schuster, Urs~M. Bergmann, and Roland
  Vollgraf.
\newblock Multivariate {{Probabilistic Time Series Forecasting}} via
  {{Conditioned Normalizing Flows}}.
\newblock In \emph{International {{Conference}} on {{Learning
  Representations}}}, September 2020.

\bibitem[Salinas et~al.(2020)Salinas, Flunkert, Gasthaus, and
  Januschowski]{Salinas2020DeepARProbabilisticForecasting}
David Salinas, Valentin Flunkert, Jan Gasthaus, and Tim Januschowski.
\newblock {{DeepAR}}: {{Probabilistic}} forecasting with autoregressive
  recurrent networks.
\newblock \emph{International Journal of Forecasting}, 36\penalty0
  (3):\penalty0 1181--1191, July 2020.
\newblock ISSN 0169-2070.
\newblock \doi{10.1016/j.ijforecast.2019.07.001}.

\bibitem[Times(2021)]{NewYorkTimes2021}
The New~York Times.
\newblock {Coronavirus (Covid-19) Data in the United States}, 2021.
\newblock URL \url{https://github.com/nytimes/covid-19-data}.

\bibitem[Wang et~al.(2021)Wang, Maddix, Faloutsos, Wang, and Yu]{wang2021}
Rui Wang, Danielle Maddix, Christos Faloutsos, Yuyang Wang, and Rose Yu.
\newblock Bridging physics-based and data-driven modeling for learning
  dynamical systems.
\newblock In Ali Jadbabaie, John Lygeros, George~J. Pappas, Pablo~A. Parrilo,
  Benjamin Recht, Claire~J. Tomlin, and Melanie~N. Zeilinger, editors,
  \emph{Proceedings of the 3rd Conference on Learning for Dynamics and
  Control}, volume 144 of \emph{Proceedings of Machine Learning Research},
  pages 385--398. PMLR, 07 -- 08 June 2021.
\newblock URL \url{https://proceedings.mlr.press/v144/wang21a.html}.

\bibitem[Wang et~al.(2019)Wang, Smola, Maddix, Gasthaus, Foster, and
  Januschowski]{Wang2019DeepFactorsForecasting}
Yuyang Wang, Alex Smola, Danielle Maddix, Jan Gasthaus, Dean Foster, and Tim
  Januschowski.
\newblock Deep {{Factors}} for {{Forecasting}}.
\newblock In \emph{International {{Conference}} on {{Machine Learning}}}, pages
  6607--6617. {PMLR}, May 2019.

\bibitem[Wu et~al.(2021)Wu, Pan, Chen, Long, Zhang, and
  Yu]{Wu2021ComprehensiveSurveyGraph}
Zonghan Wu, Shirui Pan, Fengwen Chen, Guodong Long, Chengqi Zhang, and
  Philip~S. Yu.
\newblock A {{Comprehensive Survey}} on {{Graph Neural Networks}}.
\newblock \emph{IEEE Transactions on Neural Networks and Learning Systems},
  32\penalty0 (1):\penalty0 4--24, January 2021.
\newblock ISSN 2162-237X, 2162-2388.
\newblock \doi{10.1109/TNNLS.2020.2978386}.

\bibitem[Xu et~al.(2018)Xu, Hu, Leskovec, and Jegelka]{Xu2018HowPowerfulAre}
Keyulu Xu, Weihua Hu, Jure Leskovec, and Stefanie Jegelka.
\newblock How {{Powerful}} are {{Graph Neural Networks}}?
\newblock In \emph{International {{Conference}} on {{Learning
  Representations}}}, September 2018.

\bibitem[Yu et~al.(2018)Yu, Yin, and
  Zhu]{Yu2018SpatiotemporalGraphConvolutional}
Bing Yu, Haoteng Yin, and Zhanxing Zhu.
\newblock Spatio-temporal graph convolutional networks: A deep learning
  framework for traffic forecasting.
\newblock In \emph{Proceedings of the 27th {{International Joint Conference}}
  on {{Artificial Intelligence}}}, {{IJCAI}}'18, pages 3634--3640, {Stockholm,
  Sweden}, July 2018. {AAAI Press}.
\newblock ISBN 978-0-9992411-2-7.

\end{thebibliography}
